# CONVERSATIONAL RUBERT FOR DETECTING COMPETITIVE INTERRUPTIONS IN ASR-TRANSCRIBED DIALOGUES

Dmitrii Galimzianov and Viacheslav Vyshegorodtsev

DeepSound.AI

## ABSTRACT

*Interruption in a dialogue occurs when the listener begins their speech before the current speaker finishes speaking. Interruptions can be broadly divided into two groups: cooperative (when the listener wants to support the speaker), and competitive (when the listener tries to take control of the conversation against the speaker's will). A system that automatically classifies interruptions can be used in call centers, specifically in the tasks of customer satisfaction monitoring and agent monitoring. In this study, we developed a text-based interruption classification model by preparing an in-house dataset consisting of ASR-transcribed customer support telephone dialogues in Russian. We fine-tuned Conversational RuBERT on our dataset and optimized hyperparameters, and the model performed well. With further improvements, the proposed model can be applied to automatic monitoring systems.*

## 1. INTRODUCTION

In conversation analysis, an "interruption" is defined as a situation in which the listener begins to speak without waiting until the current speaker has stopped talking. This either forces the speaker to stop or disrupts the continuity and regularity of their speech [1]. At the broadest level, interruptions can be separated into two groups: cooperative and competitive [1]. Cooperative interruption occurs when the second speaker demonstrates agreement, provides assistance, or requests clarification, while maintaining the conversation [2]. In contrast, competitive interruption occurs when the second speaker attempts to control the conversation despite the first speaker's desire to continue speaking [1]. This situation can be considered a type of conflict, disagreement, floor-taking, topic-changing, or tangentialization [2].

Automatic systems that can differentiate between types of interruptions can be applied in customer sentiment analysis [3,4], customer satisfaction monitoring [5], and agent monitoring [6–8]. Paper [9] provides an excellent example of such an application. The authors developed a model to classify the emotional states of a contact center customer into positive, neutral, and negative based on the competitiveness of speech overlaps. In addition, classifying interruptions is important for action selection algorithms in human-machine conversational systems [10].

Some researchers have considered speech overlap detection as a part of mono audio processing—a preparatory step for the following speaker diarization [11–16]. These studies are beyond the scope of our work because we use audio comprising separate channels for each speaker, which requires no additional diarization.

## 2. RELATED WORK

We identified four key aspects of the literature dedicated to the topic: definitions of classes, data representation, representation of the speakers' context, and machine learning methods. We systematized related works according to these aspects.

### 2.1. Definitions of Classes

We encountered surprising variability in the definitions of classes in the literature. The first group of papers used a simple system consisting of two or three classes. Papers [10,17,18] leveraged the binary classification of competitive and non-competitive overlaps. Further, in [19], a three-class taxonomy was used that contains interruptions, backchannels, and turn-changes. In [20], the authors also utilized a three-class taxonomy, this one consisting of backchannels, competitive (intrusive), and cooperative (non-intrusive) interruptions. However, these authors focused only on the distinction between the two types of interruptions.

The second group of papers applied a wider taxonomy. The authors of [21] elaborated five classes, namely backchannels, failed and successful interruptions, laughter, and other utterances. Paper [22] developed a seven-class taxonomy composed of backchannels, anticipated turn taking, complementary information, interruption, simultaneous start, "brouhaha" (constantly overlapping speakers), and no overlap. Moreover, in [23], the authors applied statistical analysis to ASR-transcribed meeting series and identified 10 overlap patterns. Each pattern is characterized by three binary features: gap or overlap, overtake or no overtake, and competitive or cooperative. Finally, paper [24] considered 12 turn-taking categories, including smooth switches, interruptions, and backchannels.

In [25], the authors applied a three-step process for interruption annotation. First, they determined the turn type: interruption, backchannel, or smooth turn-exchange. Second, they decided whether the interruption was successful. Third, they classified the interruption type as either cooperative (with three subtypes) or competitive (with four subtypes).

Several papers have focused on identifying backchannels [26–31]. Authors have used either a binary classification of utterances (backchannels or non-backchannels) or more elaborated taxonomies containing three [27] or four [29] classes.

### 2.2. Data Representation

We found only one study that used only the textual form of data [18]. While all other works exploit audio formats, some combine audio and textual formats [10,17,26,27,29–31]. Only two of these studies additionally explored the possibility of using textual data without audio data [10,17]. A few studies leveraged additional data features like facial expressions and head movements [20,32].

### 2.3. Representation of the Speakers' Context

In [20], the authors explored the effect of changing the duration of overlap, using a range of 0.2 to 1.0 seconds, whereas paper [21] preserved only overlaps lasting at least 0.3 seconds, and used five-second audio clips as the context. In other research, the authors of [27] investigated two options for audio clip duration: 1.5 and 2.0 seconds, while paper [29] involved experiments with audio clip lengths ranging from five to 20 seconds. In [27], the authors experimented with the number of words (five to 10), and in [18], the authors filtered out overlaps containing fewer than three words.

In [26], only the preceding context was used because the task was to predict the presence of backchannels. Paper [21] used overlaps only if the interrupter was silent for at least three seconds beforehand. The authors of [20] experimented with the presence of the preceding context, and paper [31] leveraged GRU-embeddings of K previous utterances. In [10], both preceding (0.2 seconds) and following (0.3 seconds) contexts of overlaps were used, while paper [22] included a four-second context before and after overlaps. Finally, in [18], the authors used two sentences: the sentence spoken by the interrupting listener and the preceding sentence uttered by the first speaker.

## 2.4. Machine Learning Methods

Various model architectures have been reported in the literature, from classical machine learning models such as random forest models [24], K-means [17,19], and SVM [10,20,28,32], to neural networks such as FFNN, RNN, and CNN [10,24,26,27,29,32]. Natural language processing (NLP) models include bag-of-words [17], bag-of-ngrams [10], and Word2Vec [26,27]. Recent studies used more advanced architectures, such as WavLM [21,22], BERT [29–31], and GPT [18].

## 3. MATERIALS AND METHODS

We developed a text-based model for interruption classification because such models do not receive as much attention as audio-based models (see Section 2.2). To develop our model for potential application in contact center monitoring, we used ASR-transcribed customer support telephone dialogues.

## 3.1. Data Schema

We reorganized the structure of a transcription table in CSV format that was produced by ASR and consisted of rows corresponding to VAD-separated dialogue turns. First, for each dialogue turn K containing the speaker's phrase and timestamps of its start and end, we concatenated the same information about the following dialogue turn K+1.

Second, we added four columns with the derived features. Based on the timestamps, we classified switches between speakers as a gap (if there was a pause between speakers' turns) or an overlap (if the start of turn K+1 occurred before the end of turn K). Following [2], we classified each overlap as **successful** if the end of turn K occurred before the end of turn K+1 or **unsuccessful** if the end of turn K occurred no earlier than the end of turn K+1. An unsuccessful overlap indicated that **the interrupter did not overtake a turn.** In addition, we implemented a binary feature to indicate whether a client played the role of interrupter.

Finally, we computed each overlap's duration based on the timestamps. We defined the end of an overlap as the moment when one of the speakers stopped talking. Therefore, each overlap was described by the speakers' phrases, successfulness, duration, and the client's position in the interruption.

## 3.2. Data Filtration and Labeling

The detection of competitive interruptions is very important for call center monitoring. Following [10,17,18], we leveraged a simple binary classification system involving competitive (intrusive) and non-competitive (non-intrusive) interruptions. Non-competitive interruptions included backchannels, simultaneous starts, and other types of supportive or unintended overlaps. We filtered overlaps based on the above-defined characteristics. First, we filtered out unsuccessful interruptions. Second, we excluded all overlaps lasting less than one second; the shorter the overlap, the more likely that it is a backchannel or something not of interest. Third, interruptions caused by clients and agents were preserved because both cases are of interest for call center monitoring.

The labeling process was organized as follows. An annotator listened to the part of the audio containing the overlap and marked it as non-competitive, competitive, or undefined if they could not decide between the two classes. Overlaps marked as undefined were removed from the dataset. Tables 1 and 2 present the distribution of classes across speakers and folds, respectively.

Table 1. Distribution of classes across speakers.

| Label vs. Speaker | Clients | Agents | Total |
|---|---|---|---|
| Competitive | 519 | 713 | 1232 |
| Non-competitive | 426 | 517 | 943 |
| Undefined | 268 | 278 | 546 |

Table 2. Distribution of classes across folds.

| Label vs. Fold | 0 | 1 | 2 | 3 | 4 | 5 | 6 | 7 | 8 | 9 |
|---|---|---|---|---|---|---|---|---|---|---|
| Competitive | 123 | 124 | 125 | 120 | 121 | 123 | 123 | 124 | 116 | 133 |
| Non-competitive | 93 | 93 | 94 | 96 | 98 | 95 | 95 | 96 | 97 | 86 |

## 3.3. Model and Implementation

We fine-tuned BERT [33] for several reasons. First, despite advances in NLP, we found only one study that applied the large language model to classify interruptions [18]. All three papers that utilized BERT (see Section 2) aimed to detect backchannels but not interruptions. As BERT demonstrated good performance on a similar problem, it was worth trying to adapt it to our task. Moreover, BERT is less resource-consuming than GPT-3.5 or GPT-4 [18], which is critical for potential industrial applications.

The proposed methods were implemented primarily based on the code developed in our previous work [34] since the text classification task in that study and the present study are structurally similar. In addition, the dataset was obtained from the same area and source, and it was transcribed using the same ASR system. As "Conversational RuBERT" by DeepPavlov [35,36] demonstrated high performance, we used it in the current study. We fine-tuned the model during five epochs with a batch size of 16, weight decay of 0.01, and maximum length in the tokenizer of 128, using the default optimizer and scheduler. We followed the same nine-fold cross-validation process and tested the models on the 10th fold, applying standard classification metrics such as ROC AUC and F1.

Computer Science & Information Technology (CS & IT) 71
## 4. RESULTS

We conducted three experiments to optimize the hyperparameters. In the first experiment, we checked whether the context of the interrupted speaker was informative for the model. For this purpose, we trained the model in two settings. In the first setting, the model received only the context of the interrupter as an input; in the second setting, the model's input included the phrases of both speakers [37], as in [18]. The inclusion of data regarding the interrupted speaker's context significantly improved the model's performance.

In the second experiment, the learning rate was adjusted. The best learning rate value for the task was 7e-6. Finally, we checked whether a wider context can help the model better differentiate between competitive and cooperative interruptions. For this purpose, we concatenated eight preceding turns (phrases) to the beginning of the speaker's context and eight following turns to the end of the listener's (interrupter's) context. However, this drastically reduced the model's performance.

Table 3 summarizes the basic metrics of the experiments. Extended metrics can be found in our public GitHub repository (see "Availability of Data and Materials"). Figure 1 presents the changes in loss and metrics for the validation folds while training the best model. The highest F1-macro on the test fold was 0.8404.

Table 3. Experimental results on the test fold.

| Examined Hyper Parameter | ROC AUC binary | Best Threshold | Recall macro | Precision macro | Balanced Accuracy | F1 macro |
|---|---|---|---|---|---|---|
| Experiment 1. Influence of speakers' context. The learning rate was set to 3.e-6. | | | | | | |
| Input: interrupter only | 0.8118 | 0.5249 | 0.7120 | 0.7301 | 0.7120 | 0.7151 |
| Input: both speakers | 0.8508 | 0.5303 | 0.7571 | 0.7599 | 0.7571 | 0.7582 |
| Experiment 2. Learning rate adjustment. The input contains the context of both speakers. | | | | | | |
| Lr: 1.e-6 | 0.5566 | 0.5035 | 0.5334 | 0.5348 | 0.5334 | 0.5566 |
| Lr: 3.e-6 | 0.8508 | 0.5303 | 0.7571 | 0.7599 | 0.7571 | 0.7582 |
| Lr: 5.e-6 | 0.8818 | 0.4374 | 0.8221 | 0.8483 | 0.8221 | 0.8286 |
| Lr: 7.e-6 | 0.8870 | 0.3858 | **0.8325** | **0.8671** | **0.8325** | **0.8404** |
| Lr: 9.e-6 | **0.8891** | 0.3983 | 0.8260 | 0.8517 | 0.8260 | 0.8325 |
| Experiment 3. Influence of the extended context. The learning rate was set to 7.e-6. | | | | | | |
| The extended context | 0.7049 | 0.3949 | 0.6542 | 0.6572 | 0.6542 | 0.6534 |

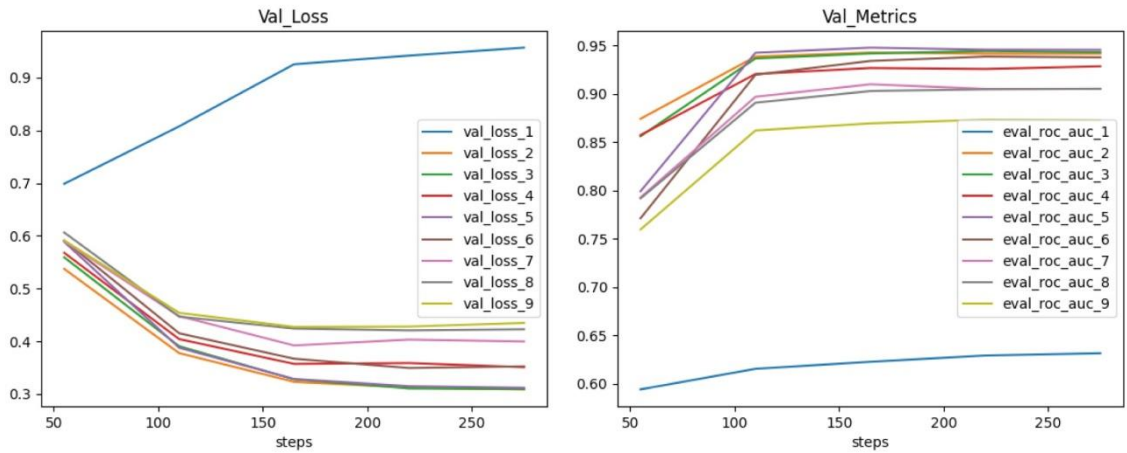

Figure 3. The training process of the best model: loss (left) and ROC AUC (right) on the validation folds.

## 5. DISCUSSION

In this section, we discuss the limitations of our work. Most of the limitations are due to the simplified labeling process; however, a more complex process would have been too resource-consuming.

- Filtration process

  - Interruptions considered as unsuccessful attempts can also indicate an interrupter's intention, whether it is aggressive or supportive. Their inclusion may enrich the model but would demand more extensive labeling.
  - The threshold value of one second used to remove too-short overlaps could be optimized experimentally.

- Labeling process

  - For better generalizability, interruptions labeled as "undefined" could be either included as a separate class or attributed to one of two existing classes.
  - One sample was labeled by one annotator only once. Such a labeling process does not allow the calculation of intra-annotator consistency or inter-annotator agreement, as was done in [22].

- Model performance

  - The developed model, despite performing well (F1-macro = 0.8404), should be optimized in order to be used in industrial applications.

## 6. CONCLUSIONS

We developed an NLP model that classifies competitive and non-competitive interruptions in dialogues. The proposed model takes as input the ASR-transcribed phrases of the speaker and the interrupting listener. From the collected dataset, we eliminated all unsuccessful interruption attempts, overlaps lasting less than one second, and overlaps that could not be unambiguously classified by annotators. The model achieved good performance, obtaining an F1-macro value of 0.8404 on the test dataset. With further improvements, the proposed model could be applied to automatic monitoring systems.

## AUTHOR CONTRIBUTIONS

Dmitrii Galimzianov: conceptualization, methodology, investigation, software, formal analysis, visualization, and writing—original draft preparation. Viacheslav Vyshegorodtsev: resources, data curation, and writing—review editing. All authors reviewed the results and approved the final version of the manuscript.

## AVAILABILITY OF DATA AND MATERIALS

The in-house dataset of transcribed calls used in this study cannot be shared due to privacy issues. The training and evaluation code is available in our public GitHub repository: https://github.com/gal-dmitry/INTERRUPTION_DETECTION_PUBLIC.

## CONFLICTS OF INTEREST

The authors declare that they have no conflicts of interest to report regarding the present study.